%% file: main.tex
\documentclass[runningheads]{llncs}

\usepackage[mobile]{eccv} 

\usepackage{eccvabbrv}

\usepackage{graphicx}
\usepackage{booktabs}
\usepackage{algorithm,algcompatible,amsmath}
\usepackage{multirow}
\usepackage{float}

\usepackage[accsupp]{axessibility}  

\usepackage{hyperref}

\usepackage{orcidlink}

\newcommand{\name}{GaussCtrl\xspace}
\newcommand\blfootnote[1]{
    \begingroup
    \renewcommand\thefootnote{}\footnote{#1}
    \addtocounter{footnote}{-1}
    \endgroup
}

\def\eg{\emph{e.g.}}
\def\ie{\emph{i.e.}}

\newcommand{\figref}[1]{Fig.~\ref{#1}}
\newcommand{\tabref}[1]{Tab.~\ref{#1}}
\newcommand{\equref}[1]{Eqn.~\ref{#1}}

\begin{document}




\title{GaussCtrl: Multi-View Consistent Text-Driven 3D Gaussian Splatting Editing}

\titlerunning{\name}

\author{Jing Wu*\inst{1}\orcidlink{0009-0007-0942-5418} \and
Jia-Wang Bian*\inst{2}\orcidlink{0000-0003-2046-3363} \and
Xinghui Li\inst{1}\orcidlink{0000-0003-3797-5082} \and Guangrun Wang\inst{1}\orcidlink{0000-0001-7760-1339} \and Ian Reid\inst{2}\orcidlink{0000-0001-7790-6423} \and Philip Torr\inst{1} \orcidlink{0009-0006-0259-5732} \and Victor Adrian Prisacariu\inst{1} 
}

\authorrunning{J. Wu et al.}

\institute{University of Oxford \and
Mohamed bin Zayed University of Artificial Intelligence \\
\email{\{jing.wu, philip.torr\}@eng.ox.ac.uk, \{jiawang.bian, ian.reid\}@mbzuai.ac.ae, \{xinghui, victor\}@robots.ox.ac.uk, wanggrun@gmail.com} 
}

\maketitle

\blfootnote{* indicates equal contribution.}
\input{00_abstract2}
\input{01_introduction2}
\input{02_related_works}
\input{03_method2}

\input{04_experiments2}
\input{05_conclusion}

\newpage

%
%
\clearpage
\bibliographystyle{splncs04}
\bibliography{egbib}

\end{document}

%% file: 00_abstract2.tex
 \begin{figure}[H]
    \centering
    \vspace{-0.8cm}
   \includegraphics[width=1\linewidth]{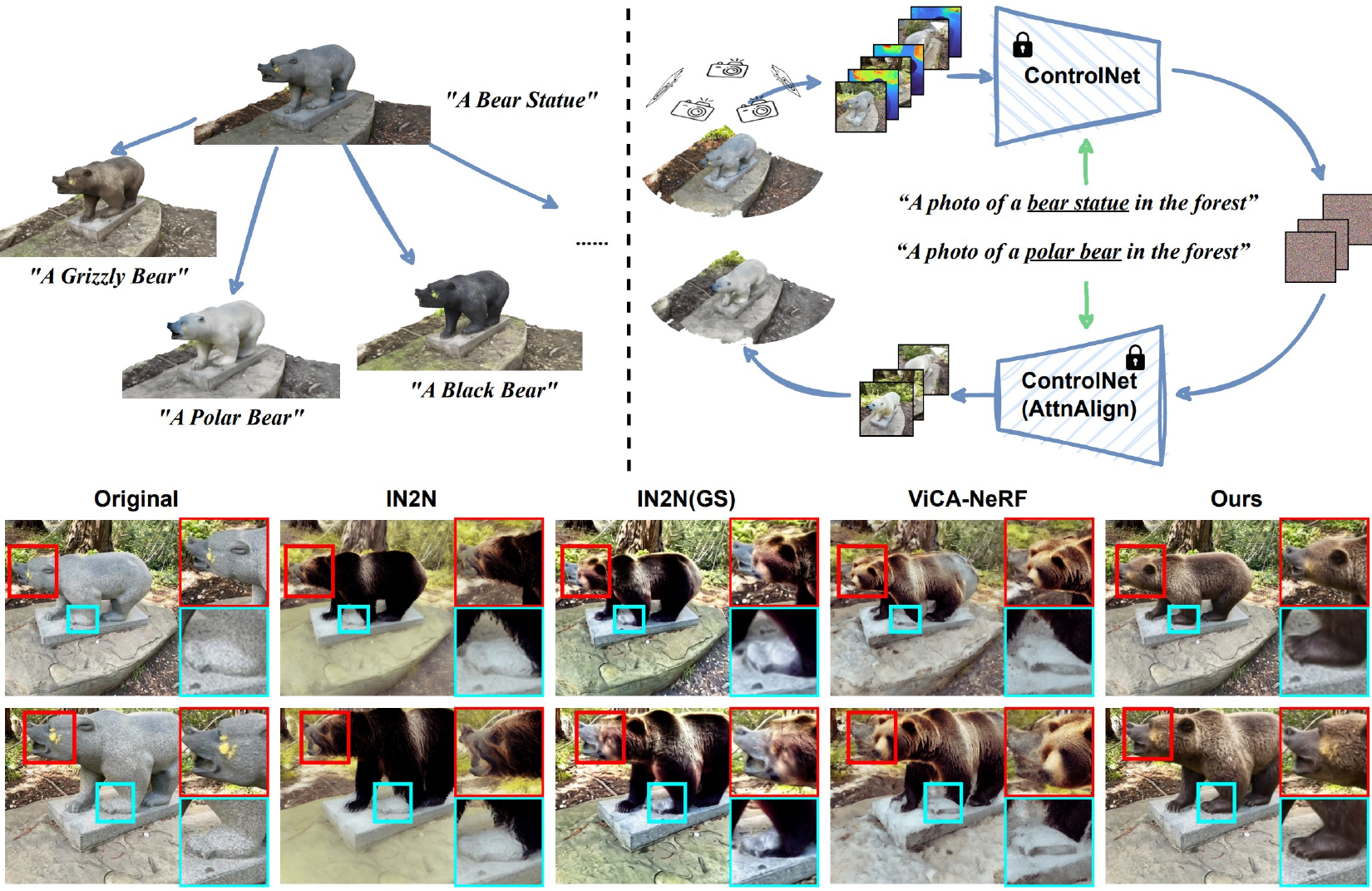}
    \caption{\name. 
    Our method edits a 3D Gaussian Splatting (3DGS) scene by modifying its descriptive prompt (Upper Left). This is achieved by editing the rendered images of 3DGS and re-training the 3D model (Upper Right). 
    Our contribution is a depth-conditioned 
    multi-view consistent editing framework, which substantially improves the blurry or unreasonable 3D results caused by inconsistent editing in previous work (Bottom). 
    } 
\label{fig:teaser}
\vspace{-0.5cm}
\end{figure}

\begin{abstract}
We propose \name, a text-driven method to edit a 3D scene reconstructed by the 3D Gaussian Splatting (3DGS).
Our method first renders a collection of images by using the 3DGS and edits them by using a pre-trained 2D diffusion model (ControlNet) based on the input prompt, which is then used to optimise the 3D model.
Our key contribution is multi-view consistent editing, which enables editing all images together instead of iteratively editing one image while updating the 3D model as in previous works.
It leads to faster editing as well as higher visual quality.
This is achieved by the two terms: 
(a) \textbf{depth-conditioned editing} that enforces geometric consistency across multi-view images by leveraging naturally consistent depth maps.
(b) \textbf{attention-based latent code alignment} that unifies the appearance of edited images by conditioning their editing to several reference views through self and cross-view attention between images' latent representations.
Experiments demonstrate that our method achieves faster editing and better visual results than previous state-of-the-art methods. 
Project website: \url{https://gaussctrl.active.vision/}
  
\keywords{3D Editing \and Diffusion Models \and Gaussian Splatting \and Neural Radiance Fields}
\end{abstract}

%% file: 01_introduction2.tex
\section{Introduction}
\label{sec:intro}

Neural representations such as Neural Radiance Field (NeRF) \cite{mildenhall2020nerf} and 3D Gaussian Splatting (3DGS) \cite{kerbl3Dgaussians} have demonstrated an ability to create a 3D reconstruction that can guarantee remarkably high-quality novel view rendering.
However, the ability to edit these resulting 3D representations for the creation of 3D assets remains a crucial yet underexplored aspect. 
The first pipeline was recently proposed by Instruct NeRF2NeRF \cite{instructnerf2023} (IN2N) to edit a NeRF scene using text-based instructions.
It leverages an image-conditioned diffusion model (Instruct Pix2Pix \cite{brooks2022instructpix2pix}) to iteratively edit the images rendered from the NeRF while updating the underlying NeRF-based reconstruction. 
Follow-ups \cite{dong2023vica, igs2gs} follow the same idea of editing 3D from 2D and show encouraging results. 

While IN2N transforms the 3D editing problem into a more manageable 2D editing task, consistency across multiple views is not encouraged. 
Such consistency is critical to the final 3D editing to prevent visual artefacts such as blurring and inconsistent appearance at different viewpoints.
However, this is challenging for modern 2D diffusion models to achieve since their input is a single image and they do not enforce geometric consistency across different views.
IN2N edits one image at a time while optimising the 3D NeRF, generating an averaged result.
This process leads to some geometric consistency, however, the convergence is slow, and it requires many full NeRF optimisations.
Although its follow-ups yield better visual quality with their own innovations, inconsistencies across rendered views remain---See \figref{fig:teaser} (Bottom). 



We propose \name to address the challenge of inconsistency.
Our method edits 3D scenes with text instructions, as illustrated in \figref{fig:teaser} (Upper Left \& Right).
Distinguishing from IN2N, we propose a depth-controlled editing framework and a latent code alignment module to explicitly enforce the multi-view consistency, enabling editing all images together and updating the 3D model only once.
Specifically,
\textbf{first}, 
as depth maps of the scene are naturally geometry-consistent, we condition image editing on them by employing ControlNet~\cite{zhang2023controlnet}. 
Images are inverted to their respective latent codes together with their depth maps via DDIM inversion \cite{song2020ddim} and then are edited by the denoising process with edited prompts. 
By doing so, the edited images inherit the consistency from depth maps, avoiding the abrupt and incoherent geometric changes that may present in other editing methods \cite{mokady2022null, hertz2022prompt2prompt}.
\textbf{Second},
we propose an attention-based latent code alignment module to encourage appearance consistency during the editing process.
Specifically, we choose several reference views and align all other views' latent codes to these reference views during the denoising process, through the use of both self and cross-view attention.
This greatly improves high-level semantic consistency over the entire dataset. 

We evaluate our approach on a variety of scenes with different text prompts, ranging from forward-facing scenes to challenging 360-degree object-centred scenes. We also perform an ablation study on different components of our method to validate their effectiveness. Experiments demonstrate that our method significantly improves the visual quality of editing and greatly reduces processing time. 
We summarise our contribution as follows:
\begin{enumerate}

    \item We propose \name, to enable efficient editing of 3DGS scenes with text instructions.
    
    \item We employ depth guidance and the attention-based latent code alignment module to encourage multi-view consistent editing.

    \item The proposed method demonstrates more realistic editing and achieves higher visual quality than previous work
    on a variety of 3D editing scenes.

\end{enumerate}

%% file: 02_related_works.tex

\section{Related Works}

\subsection{2D Editing with Diffusion Models}


Diffusion models \cite{song2020ddim, ho2020ddpm, Nichol2021ImprovedDD, Luo2022UnderstandingDM, Sohl2015diffusion, hu2022lora} have gained popularity in image generation due to their ability to produce highly realistic images. 
By training on billions of image-text pairs, these models not only offer the flexibility to customise generation through textual prompts \cite{rombach2022high, Dhariwal2021DiffusionMB, Ho2022ClassifierFreeDG, Saharia2022imagen, Ramesh2022Dalle} but also enable various forms of editing. 
Most editing methods leverage pre-trained Stable Diffusion \cite{rombach2022high}. Starting with inverting the latent representation of the to-be-edited image to its corresponding noise by DDIM inversion, the editing is achieved through the denoising process. One form of editing \cite{pan2023draggan, mou2023dragondiffusion, mou2023diffeditor, epstein2023selfguidance} allows users to label several anchor points and drag them to target positions. 
DragDiffusion \cite{shi2023dragdiffusion} optimises the latent code of an image by minimising feature differences between initial and target positions. 
SDE-Drag \cite{nie2023sdedrag} replaces the optimisation with copy-and-paste of features by switching to DDPM scheduler \cite{ho2020ddpm}. 
Another form is to edit images through textual prompts, which is more relevant to our work. Delta denoising score (DDS) \cite{hertz2023dds} directly optimises the image latent representation by forcing the similarity between noises predicted using the original and edited texts. Prompt-to-Prompt (P2P) \cite{hertz2022prompt2prompt} achieves editing by manipulating the cross attention between image and text. Null-text inversion \cite{mokady2022null} tackles artefacts at DDIM inversion when using classifier-free guidance \cite{Ho2022ClassifierFreeDG} and integrates their method to P2P. 
While significant progress has been made in 2D image editing, none of them considers multi-view consistency, which may lead to artefacts in editing. 
Our method, tailored for 3D editing, ensures such consistency through initial latent code control and newly proposed attention-based latent code alignment.

\subsection{3D Editing in NeRF and Gaussian Splatting}

NeRF and 3DGS are two of the most popular 3D models for neural novel view synthesis. NeRF implicitly encodes the geometry and colours of a scene in a Multi-Layer Perceptron (MLP), whereas 3DGS explicitly expresses the scene as Gaussian ellipse point clouds. Although they exhibit promising results in 3D reconstruction, their editing remains challenging. Current attempts can be largely categorised into two main categories: 3D Style Transfer, and Text-driven Editing. 


\noindent \textbf{3D Style Transfer: } Similar to 2D style transfer \cite{Gatys2016styletransfer}, 3D style transfer aligns the style of a 3D scene to the style of a provided reference image. Notable examples include StyleRF \cite{liu2023stylerf}, StylizedNeRF \cite{Huang22StylizedNeRF}, ARF \cite{zhang2022arf}, and PaletteNeRF \cite{li2022palettenerf}. However, this line of work fails to edit the local details of the scene, and the reference image is not always available.


\noindent \textbf{Text-driven Editing: } Instruct NeRF2NeRF (IN2N) \cite{instructnerf2023} is the first NeRF editing work that edits 3D models with text instructions. 
This method effectively transforms the 3D editing challenge into a 2D image editing task. 
By rendering images from the 3D scene and editing them using Instruct Pix2Pix (IPix2Pix) \cite{brooks2022instructpix2pix}, IN2N iteratively updates the 3D scene until convergence. 
As there is no guarantee of consistent editing of multi-view images, this method suffers from instability, slow processing speeds, and notable artefacts, particularly evident in 360-degree scenes.
ViCA-NeRF \cite{dong2023vica}, following a similar idea to IN2N, selects several reference images from the dataset of the scene, edits them by IPix2Pix, and edits the rest of the dataset as blended results of the projection of reference images to alleviate the inconsistency. 
However, the blending does not fully address the consistency issue and suffers from blurry editing. 
DreamEditor \cite{zhuang2023dreameditor} converts NeRF to mesh and directly optimises the mesh with SDS loss \cite{poole2022dreamfusion} and DreamBooth \cite{ruiz2023dreambooth}. 
Our method shares similarities with IN2N and ViCA-NeRF but aims to address their limitations and offer superior consistency and visual quality in the editing results.

 \begin{figure}[t]
\centering
   \includegraphics[width=1\linewidth]{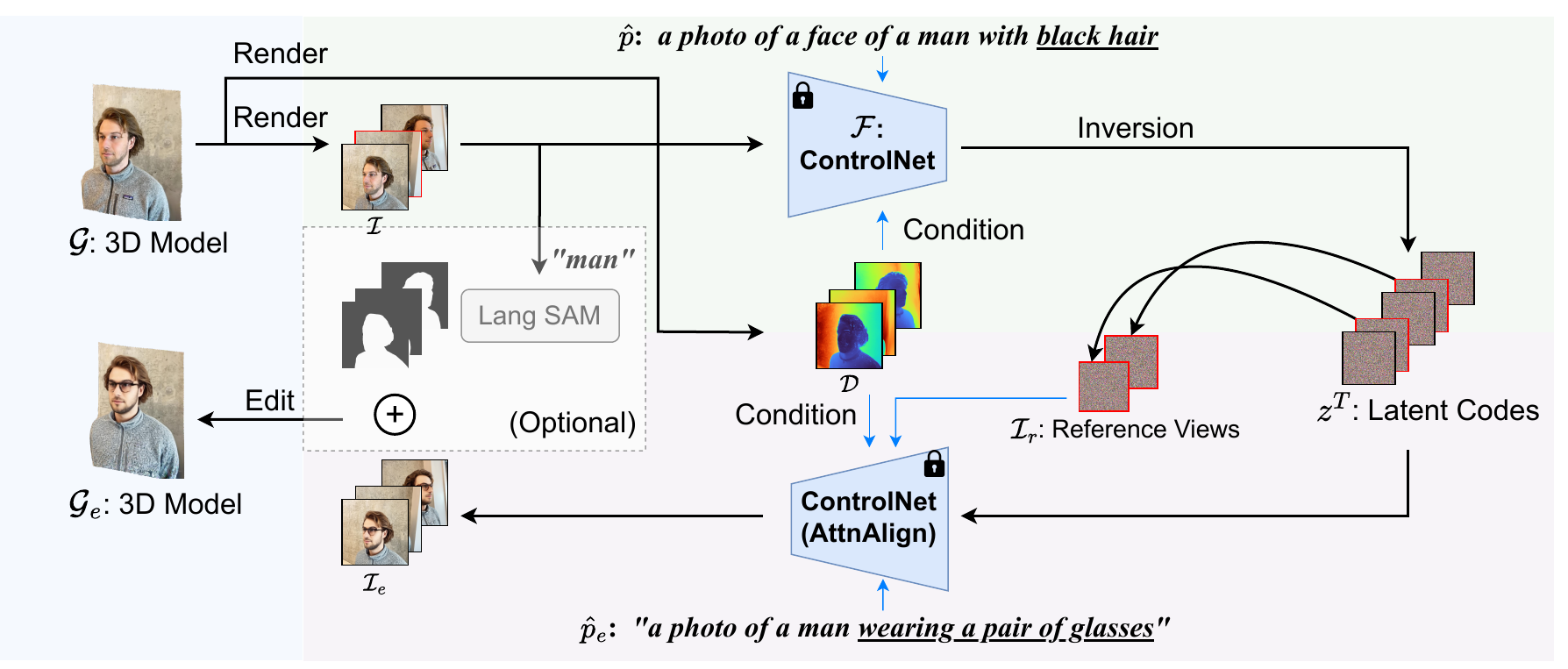}
    \caption{\name pipeline. Given a 3DGS scene and text instructions, our method renders images using the 3DGS and edits the rendered images with text instructions, which are then used to optimise the original 3DGS. Our key contribution is multi-view consistent editing. Towards this, we propose (1) depth-conditioned editing based on ControlNet for geometry consistency; and (2) attention-based latent code alignment for improving consistency during editing.
    } 
\label{fig: pipeline}
\vspace{-0.5cm}
\end{figure}

%% file: 03_method2.tex
\section{Method}

We propose \name, a novel approach to edit a 3D Gaussian Splatting (3DGS) model using textual prompts. 
Given a collection of images and their reconstructed 3D model, our method first re-renders each dataset image to the required resolution and renders their respective depth maps. 
Then, we employ ControlNet~\cite{zhang2023controlnet} to conduct depth-conditioned editing for all images supplemented by attention-based latent code alignment to encourage geometry and appearance consistency. 
Finally, we optimise the original 3D model using the edited images to obtain the new edited 3D model. 
Optionally, a mask generated by Language-based Segment Anything (Lang SAM) \cite{kirillov2023sam} is applied to filter the background for better quality when editing local objects.
The comprehensive pipeline is illustrated in \cref{fig: pipeline}. 
In the following, we commence by reviewing the background of 3DGS and ControlNet in \cref{sec:3.1 background}, followed by the introduction of our proposed methods, including depth-conditioned image editing in \cref{sec:depth-editing}, and attention-based latent code alignment in \cref{sec: Cross-View Attention Alignment Consistency Control}.

\subsection{Background}
\label{sec:3.1 background}

\noindent \textbf{3D Gaussian Splatting: }
Gaussian Splatting \cite{kerbl3Dgaussians} is an explicit 3D representation based on point clouds. A set of 3D Gaussians is modelled to represent the scene. Each Gaussian ellipse has a colour and an opacity and is defined by its centred position $x$ (mean), and a full covariance matrix $\Sigma$:
$G (x) = e^{-\frac{1}{2} x^T \Sigma^{-1} x}$. When projecting 3D Gaussians to 2D for rendering, a method of splatting \cite{Zwicker2001splatting} is used to position the Gaussians on 2D planes, which involves a new covariance matrix $\Sigma'$ in camera coordinates defined as $\Sigma' = J W \Sigma W^T J^T$, where $W$ denotes a given viewing transformation matrix and $J$ is the Jacobian of the affine approximation of the projective transformation. To enable differentiable optimisation, $\Sigma$ is further decomposed into s scaling matrix $S$ and a rotation matrix $R$: $\Sigma = RSS^TR^T$.


\noindent \textbf{ControlNet: } ControlNet \cite{zhang2023controlnet} is an end-to-end spatial conditional generation model built on top of Stable Diffusion (SD) \cite{rombach2022high}. It implants additional U-Net encoders to SD, which enables image generation controlled by various kinds of extra information, \eg, depth, normals, edges, or hand-drawn priors. We employ ControlNet for its ability of depth-controlled generation. 

\subsection{Depth-conditioned Image Editing}\label{sec:depth-editing}

Previous work~\cite{instructnerf2023} employs Instruct-Pix2Pix~\cite{brooks2022instructpix2pix} for image editing, resulting in visually appealing individual images.
However, ensuring multi-view consistency between these images remains a challenge, often resulting in visual artefacts and unstable editing outcomes.
To this end, we conduct depth-conditioned image editing by employing ControlNet $\mathcal{F}$, comprising a U-Net block $\mathcal{F}_U$ and a ControlNet block $\mathcal{F}_C$. 
As the depths $\mathcal{D}$ are extracted from the 3D model, they are naturally consistent across multiple views.
By conditioning image editing on these consistent depth maps, our method effectively promotes consistency in 3D geometry across all edited images.

Given a to-be-edited image $\mathcal{I}$ and its corresponding description prompt $\hat{p}$, we begin by computing its latent code $z^0$, using the VAE encoder of the ControlNet. We then iteratively invert it to its corresponding Gaussian noise $z^T$ via DDIM inversion. Mathematically, the inversion can be described as follows: 
\begin{align}
  \epsilon^{t} &= \mathcal{F}_U(z^t, t, \hat{p}, \mathcal{F}_C(z^t, t, \hat{p}, \mathcal{D})) \\
  z^{t+1} &= \sqrt{\alpha_{t+1}} \frac{z^t - \sqrt{1-\alpha_t } \cdot \epsilon^t}{\sqrt{\alpha_t}} + \sqrt{1 - \alpha_{t+1}} \epsilon^t
  \label{eq:ddim inv}
\end{align}
where $t$ is the time step of the diffusion process and $\alpha_{t}$ is the scheduling coefficient in DDIM scheduler.
After reaching $z^T$, we replace the original prompt $\hat{p}$ with the edited prompt $\hat{p}_e$ containing changed content, and obtain the edited latent code $z^{0}_{e}$ through the denoising process:
\begin{align}
  \epsilon_{p}^{t} & = \mathcal{F}_U(z_e^t, t, \hat{p}_e, \mathcal{F}_C(z_e^t, t, \hat{p}_e, \mathcal{D})) \label{eq: cond_noise} \\
  \epsilon_{\emptyset}^{t} & = \mathcal{F}_U(z_e^t, t, \emptyset, \mathcal{F}_C(z_e^t, t, \emptyset, \mathcal{D})) \label{eq: uncond_noise} \\
  \epsilon^{t} &= \epsilon_{\emptyset}^{t} + \omega \cdot (\epsilon_{p}^{t}-\epsilon_{\emptyset}^{t}) \label{eq: classifier free guidance} \\
  z_e^{t-1} & = \sqrt{\alpha_{t-1}} \frac{z_e^t - \sqrt{1-\alpha_t }\cdot \epsilon^t}{\sqrt{\alpha_t}} + \sqrt{1 - \alpha_{t-1}} \epsilon^t
  \label{eq:depth image editing}
\end{align}
where $z_e^t$ denotes the latent code of the edited images ($z_e^T = z^T$), $\emptyset$ is an empty prompt and \cref{eq: classifier free guidance} is the classifier-free guidance \cite{Ho2022ClassifierFreeDG} to improve the fidelity of the editing to the edited prompt $\hat{p}_e$. 
We obtain the final image $\mathcal{I}_{e}$ by decoding $z_e^0$ using the VAE decoder of the ControlNet.

\noindent \textbf{Discussion on DDIM inversion:}
The original ControlNet operates as a generative model, typically accepting randomly initialised noise $z^T$ as input, thus yielding diverse results. 
However, for editing tasks, we adopt a different approach by reversing the to-be-edited images into noise and utilising them as input for ControlNet. 
By doing so, the output is conditioned on the original image, and more importantly, multi-view consistency is improved during editing. 
This is because the original images have naturally consistent colour and geometry,
where we use DDIM inversion to obtain consistent initial noises for all to-be-edited images towards consistent editing.


\subsection{Attention-based Latent Code Alignment}
\label{sec: Cross-View Attention Alignment Consistency Control}
While our depth-conditioned editing approach enhances geometric consistency, individual images are still edited independently, posing challenges for appearance consistency. 
Despite sharing the same editing prompt, edited images may exhibit discrepancies in colours or produce peculiar results, particularly at challenging viewpoints. 
Previous studies \cite{hertz2022prompt2prompt, cao2023masactrl} have identified a relationship between the appearance of images generated by diffusion models and the key-value pairs in the image self-attention mechanism of the U-Net.
Inspired by this fact, we propose an attention-based latent code alignment module that explicitly aligns the appearance of images to selected reference views during editing. 
Therefore, images are no longer edited independently; instead, their appearances are unified to a common standard. 
This ensures greater consistency across edited images and mitigates issues related to appearance discrepancies.

Specifically, we first define the attention between two latent codes $z_i$ and $z_j$ as:
\begin{align}
\vspace{-0.5cm}
  \text{Attn}_{i,j} & = \text{Softmax}(\frac{W_q(z_i) W_k(z_j)^\top}{\sqrt{c}})W_v(z_j),
  \label{eq: cross-atten}
\end{align}
where $W_q(\cdot)$, $W_k(\cdot)$, and $W_v(\cdot)$ are linear projections to obtain query, key, and value for the attention operation, and $c$ is a scaling factor.
Given latent codes of $N_r$ reference images $z^t_{r, i}$, where $i=1,2,...,N_r$ and the latent code of the to-be-edited image $z^t_e$ at the time step $t$, our alignment module is defined as:
\begin{align}
\vspace{-0.5cm}
  \text{AttnAlign}_{e} & = 
  \lambda \cdot \text{Attn}_{e,e} + {(1-\lambda) \cdot \frac{1}{N_{r}} \sum_{i=1}^{N_{r}} \text{Attn}_{e,i}}
  \label{eqn:attnalign}
\end{align}
where $\lambda\in [0, 1]$. 
This module blends the self-attention of $z^t_e$ with the cross-view attention between $z^t_e$ and each reference view $z^t_{r, i}$.
The cross-view attention aligns the appearance of all edited images to the reference views while the self-attention helps each edited image retain its distinctiveness. 
We find that this design significantly improves the appearance consistency and minimises anomalies at challenging viewpoints. 
We replace all image self-attention modules with the proposed alignment module in the U-Net block $\mathcal{F}_U$ and the ControlNet block $\mathcal{F}_C$. Therefore, \cref{eq: cond_noise} and \cref{eq: uncond_noise} becomes:
\begin{align}
  \epsilon_{p}^{t} & = \mathcal{F}_U(z_e^t, z_r^t, t, \hat{p}_e, \mathcal{F}_C(z_e^t, z_r^t, t, \hat{p}_e, \mathcal{D})) \\
  \epsilon_{\emptyset}^{t} & = \mathcal{F}_U(z_e^t, z_r^t, t, \emptyset, \mathcal{F}_C(z_e^t, z_r^t, t, \emptyset, \mathcal{D}))
\end{align}

%% file: 04_experiments2.tex
\section{Experiments}

In this section, we first introduce the setup of experiments, including the testing data, method baselines, evaluation metrics, and implementation details. We then provide both qualitative and quantitative results of our method, followed by ablation studies on each proposed component and a limitation discussion.

\subsection{Experiment Setup}
\noindent\textbf{Dataset:}
To validate the effectiveness of \name, we collect a variety of scenes from multiple existing datasets for evaluation.
Specifically, we collected four 360-degree scenes from IN2N~\cite{instructnerf2023}, Mip-NeRF360~\cite{barron2022mipnerf360}, and BlendedMVS~\cite{yao2020blendedmvs} datasets, and two forward-facing scenes from IN2N~\cite{instructnerf2023} and NeRF-Art \cite{wang2022nerfart} dataset.
For each scene, we evaluate our method on multiple text-based instructions.

\noindent\textbf{Baselines:}
We mainly compare \name with two state-of-the-art methods: Instruct-GS2GS \cite{igs2gs}, a very recent update of Instruct NeRF2NeRF (IN2N) \cite{instructnerf2023} that replaces the NeRF in IN2N with a 3DGS model and ViCA-NeRF~\cite{dong2023vica} for the similarities shared between them. 
We denote Instruct-GS2GS as IN2N(GS) in the following paragraphs for it being a variant of IN2N. 
Both IN2N(GS) and VICA-NeRF employ Instruct-Pix2Pix, which takes an instruction-like prompt to edit images. Our method is based on Stable Diffusion, using a description-like prompt. Therefore, to edit a scene, we use editing instructions in IN2N(GS) and VICA-NeRF while modifying the original scene description in our method.
To ensure a fair comparison, we preprocess all dataset images to $512\times512$ resolution and evaluate all methods on the same data. 
We provide more visual comparison results, including a comparison with IN2N in our supplementary.

\begin{figure}[t]
\centering
   \includegraphics[width=.95\linewidth]{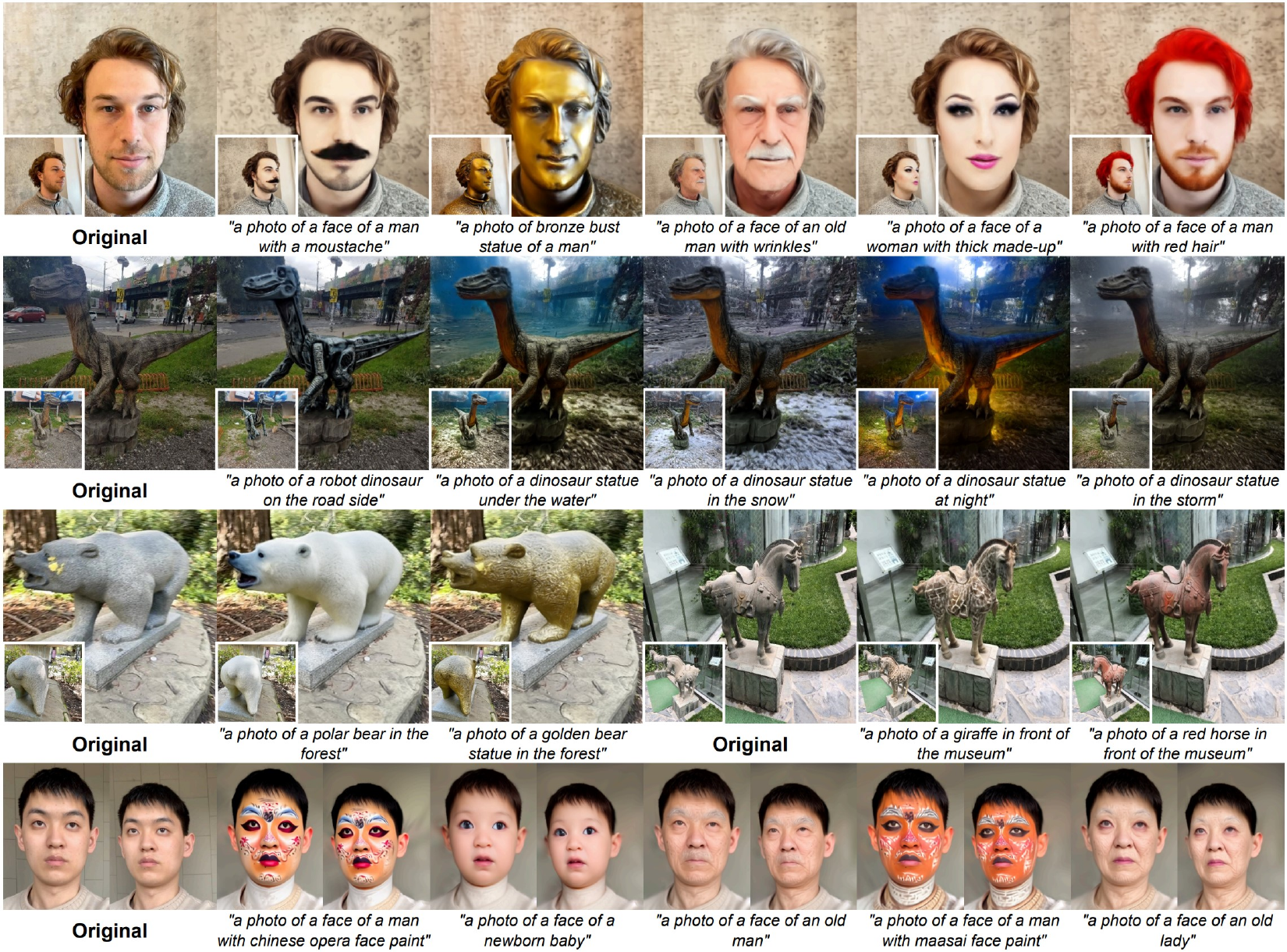}
    \caption{Qualitative results. We show diverse results of text-guided editing in various scenes, ranging from editing objects to adjusting environments, \eg, changing the appearance and age of the target human, and modifying the environment. 
    } 
    \label{fig:all_results}
    \vspace{-0.5cm}
\end{figure}

\noindent\textbf{Evaluation Criteria: } 
Following previous methods \cite{instructnerf2023, gal2021stylegannada, brooks2022instructpix2pix}, we use CLIP Text-Image Directional Similarity ($\text{CLIP}_{dir}$) to evaluate the alignment of the 3D edit with text instructions. However, we notice that this metric may not always reflect the true visual quality of editing, on which we will elaborate in later paragraphs. 

\begin{figure}[t]
\centering
   \includegraphics[width=1\linewidth]{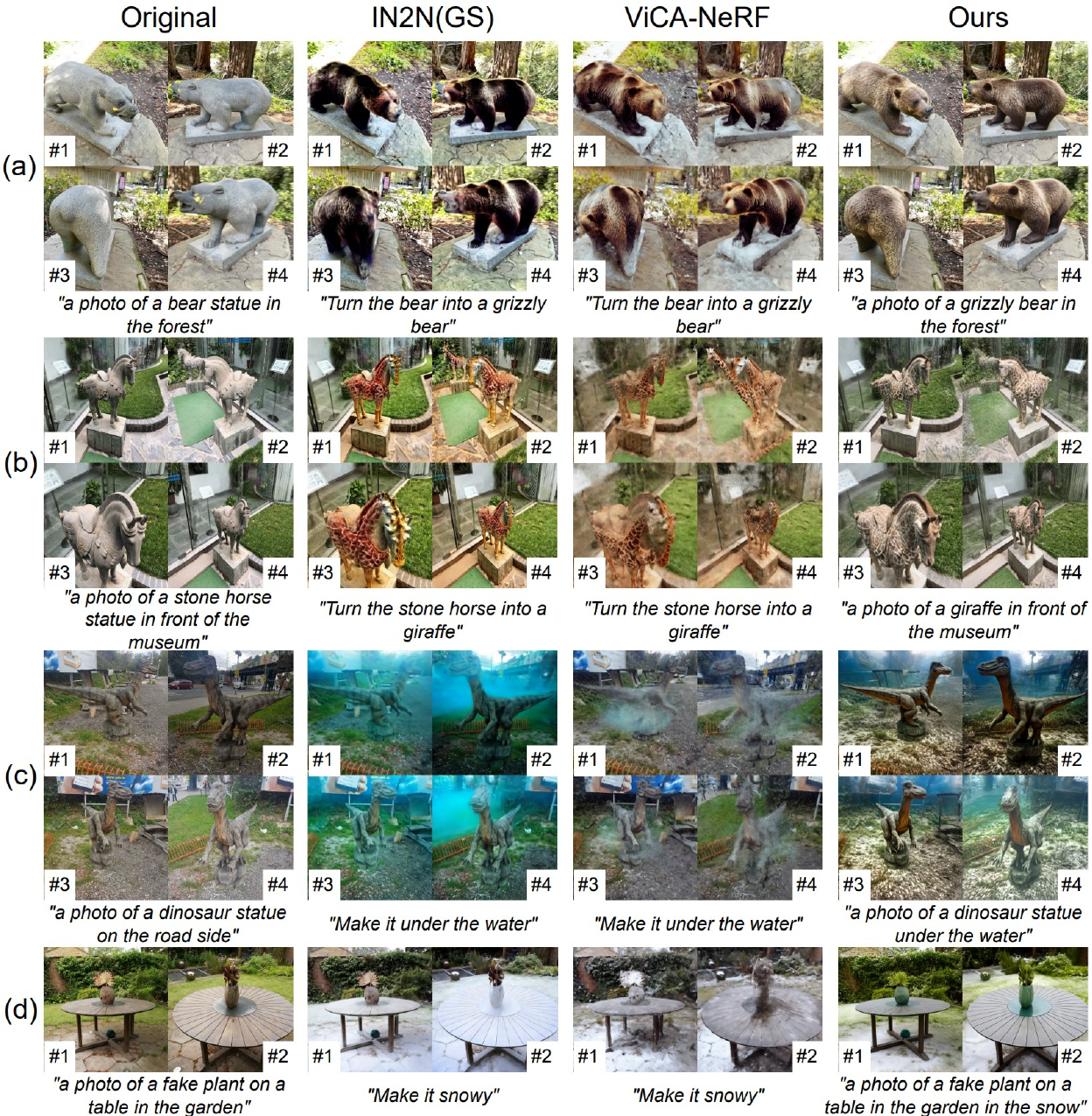}
    \caption{Qualitative comparison on 360-degree scenes. Our method generates more consistent and higher-quality images than previous state-of-the-art methods.} 
    \label{fig:qualitative_360}
    \vspace{-0.5cm}
\end{figure}

\noindent\textbf{Implementation Details: }
Our method is implemented by using the PyTorch library. 
We use the "splatfacto" model, an improved implementation of 3D Gaussian Splatting, in NeRFStudio \cite{nerfstudio} library for 3D reconstruction.
We employ Stable Diffusion v1.5 and its corresponding ControlNet for 2D image editing using the Huggingface library \cite{von-platen-etal-2022-diffusers}.
For reference views, we sample $N_r=4$ views from the dataset images randomly. 
We set $\lambda$ in \equref{eqn:attnalign} as 0.6.
We apply Language-based Segment Anything (Lang SAM) \cite{kirillov2023sam, zhang2022dino, liu2023groundingDINO, li2022dn, liu2022dabdetr} as our mask segmentation backbone. 
Our method takes around 9 minutes to edit one scene on an NVIDIA RTX A5000 with a GPU memory of 24GB. 

\begin{figure}[t]
\centering
   \includegraphics[width=0.95\linewidth]{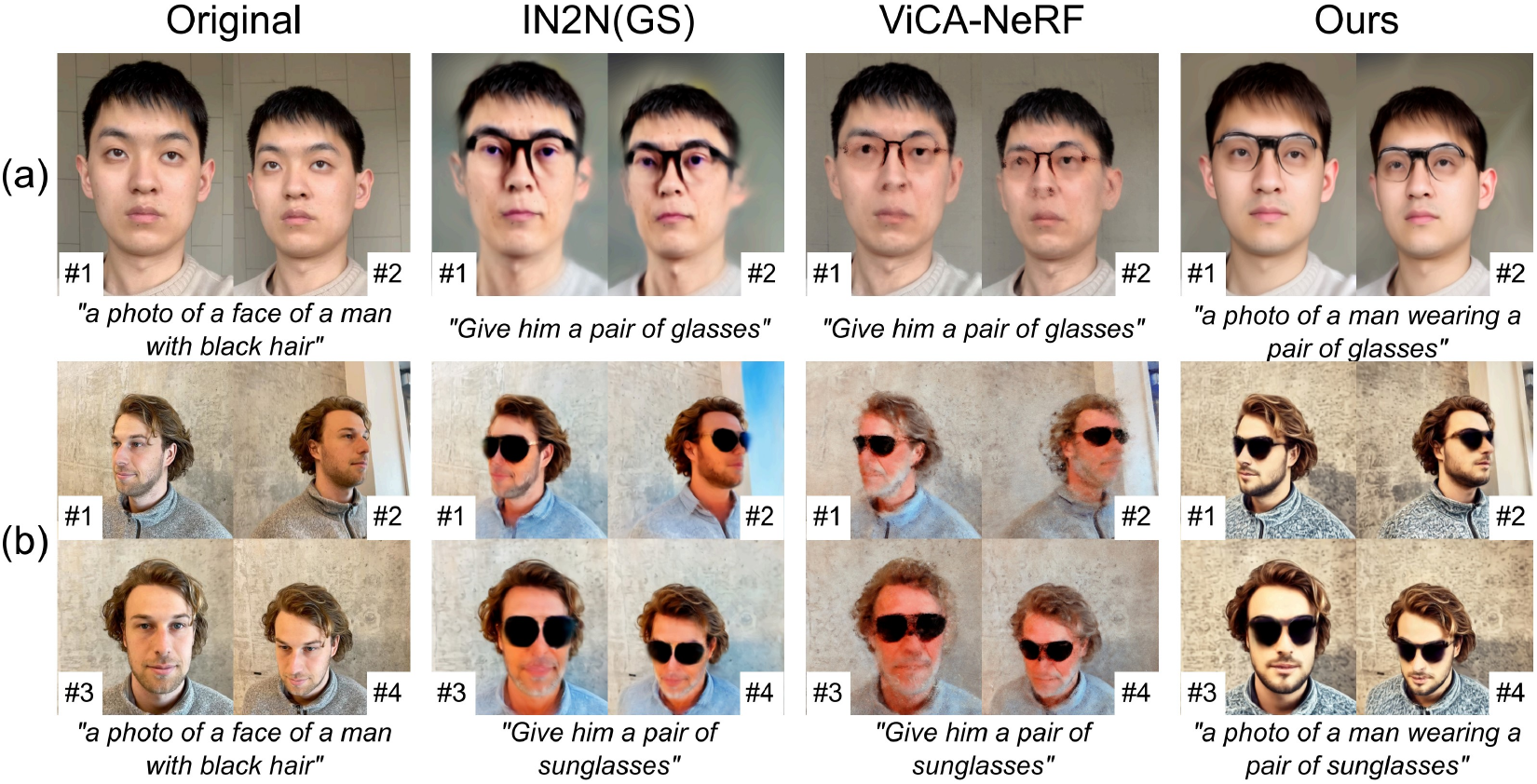}
    \caption{Qualitative results on forward-facing scenes. Our method generates more realistic results with better quality, consistency, and less artefact. 
    } 
    \label{fig:qualitative_face_forward}
    \vspace{-0.5cm}
\end{figure}

\subsection{Qualitative Evaluation}

\begin{figure}[t]
\centering
   \includegraphics[width=.85\linewidth]{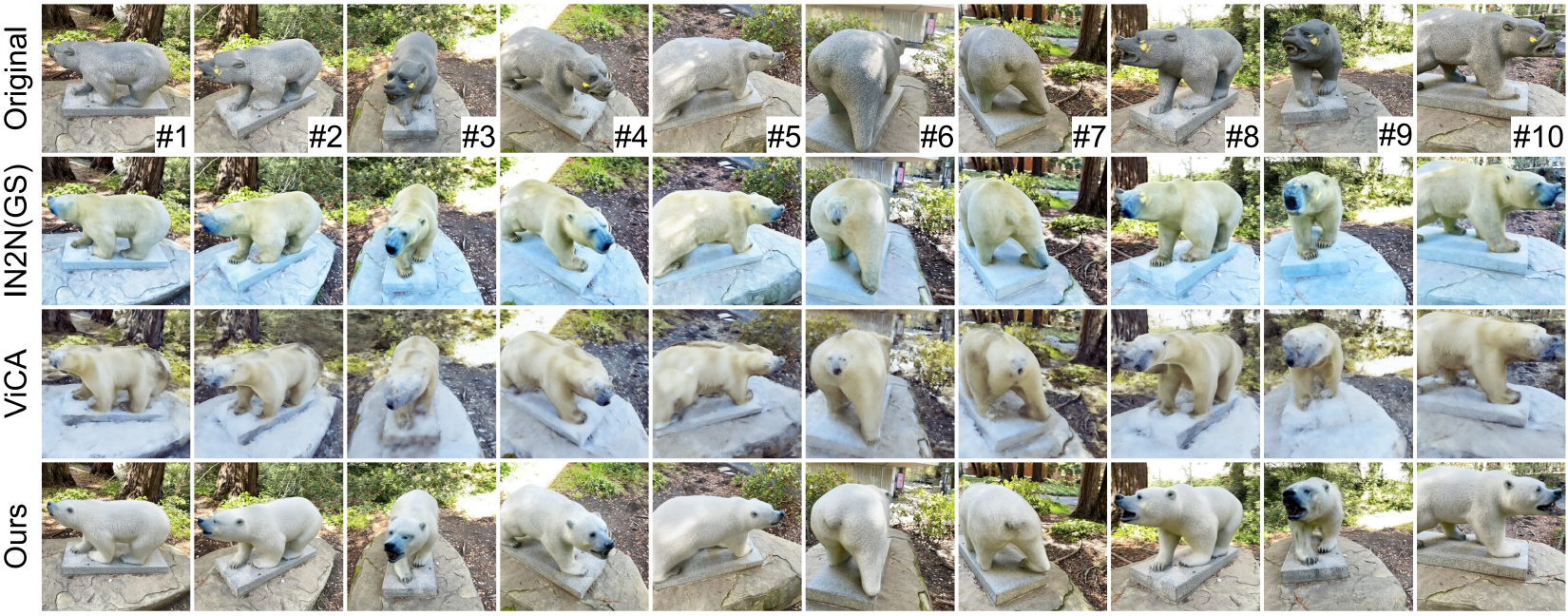}
    \caption{Editing consistency comparison on the bear scene (Polar bear). IN2N(GS) and ViCA both suffer from editing inconsistency (View \#1,2,4,8,10), which results in artefacts and blurry of the bear's face. Additionally, they are affected by the face-on-the-back problem. Our method improves on both problems. 
    } 
    \label{fig:qualitative_consistency}
    \vspace{-0.5cm}
\end{figure}

\figref{fig:all_results} illustrates various editing results of our method, including in both 360-degree and forward-facing scenes.
The text instructions range from editing objects to adjusting environments.
\figref{fig:qualitative_360} and \figref{fig:qualitative_face_forward} show the qualitative comparison of our method with previous SOTA alternatives in 360-degree and forward-facing scenes, respectively.
In \figref{fig:qualitative_consistency}, we sample $10$ different views of a scene to visually compare the editing consistency of different methods.
We make more detailed analyses below.

\figref{fig:all_results} shows that our method can perform realistic edits globally (\textit{"a photo of a dinosaur \underline{in the snow}"}) and locally (\textit{"a photo of a \underline{polar bear} in the forest"}) with high quality and consistency.
Also, it can change the colour of a specific area (\textit{"a photo of the face of a man with \underline{red hair}"}) and texture of the object (\textit{"a photo of a \underline{bronze bust statue} of a man"}).
Our method also shows good consistency for complicated texture editing such as \textit{"a photo of a face of a man with \underline{Chinese opera face paint}"}.

\figref{fig:qualitative_360} shows qualitative comparisons between our method and baselines in 360-degree scenes. 
Specifically, when editing objects, IN2N(GS) suffers from incomplete editing shown by the face area indicated by view \#2,4, the feet area in view \#1,2,4 of scene (a) and the neck area in view \# 1, 3 of scene (b). 
These show the inconsistent editing from using Instruct Pix2Pix when editing local objects. 
When editing scene's environments, IN2N(GS) demonstrates relatively better quality than editing objects as shown by scenes (c) and (d)
VICA-NeRF suffers from blurry results in both local object and scene environment editing. 
We attribute these artefacts to their blending of inconsistently edited reference images. 
Thanks to the depth-conditioned editing and proposed latent code alignment module, 
our method demonstrates sharper, more consistent and realistic results in both object and environment editing, indicating the superiority of our method over previous alternatives.

\figref{fig:qualitative_face_forward} shows qualitative comparisons in forward-facing scenes.
Compared with 360-degree scenes, IN2N(GS) and ViCA-NeRF show better consistency in the forward-facing setting. 
This is attributed to that the variation of image viewpoints in the forward-facing setting is not as extreme as in the 360-degree setting, meaning that individual image editing may retain certain consistency. 
However, compared with our method, IN2N(GS) and ViCA-NeRF still suffer from artefacts such as blurry boundaries. 
Our method also generates more realistic results.

To highlight the improvement in multi-view consistency, we render $10$ views from different angles around the object for each method in \figref{fig:qualitative_consistency}.
As shown in view \#1,2,4,8,10, both IN2N(GS) and ViCA-NeRF lose many details and are blurry on side views of the polar bear, a direct result of inconsistent editing.
Second, ViCA-NeRF loses most of the details of the bear's face as indicated in view \#1,2,3,8,9,10.
What's more, it can be observed in view \#6 that IN2N(GS) and ViCA both suffer from the face-on-the-back problem, which is caused by Instruct Pix2Pix forcefully producing a polar bear and fitting it to the image layout. 
Our approach greatly mitigates this problem through our latent code alignment module by conditioning editing on reference views. 
More detailed comparisons of editing consistency and quality are included in the supplementary material. 

\begin{table}[tb]
  \scriptsize	
  \caption{Quantitative Evaluation. $\text{CLIP}_{dir}$: CLIP Text-Image Direction Similarity
  }
  \label{tab:quantitative eval}
  \centering
  \begin{tabular}{c|l|cc|cc|cc|cc}
    \hline 

    \hline
    & \multirow{2}*{Scene}  & \multicolumn{2}{c|}{IN2N} & \multicolumn{2}{c|}{IN2N(GS)} & \multicolumn{2}{c|}{ViCA-NeRF} & \multicolumn{2}{c}{\textbf{Ours}}\\
    \cline{3-10}
    &  & $\text{CLIP}_{dir}$  & Time & $\text{CLIP}_{dir}$ & Time & $\text{CLIP}_{dir}$ & Time & $\text{CLIP}_{dir}$ & Time \\
      \hline
   \multirow{4}*{360} & Bear Statue  & 0.1019  &  \multirow{6}*{$\sim$1.5h}  & 0.1165   & \multirow{6}*{$\sim$13.5min}  &  0.1104  & \multirow{6}*{$\sim$38.5min} &  \textbf{0.1388}  & \multirow{6}*{$\sim$9min} \\
    & Dinosaur & 0.1466  &   & 0.1490   &   &  0.0723  &  &  \textbf{0.1584}  &  \\
     & Garden  & \textbf{0.3027}  &    & 0.1663 &   &  0.2903  & & 0.2891  &  \\
    & Stone Horse  &  0.1654 &    &  0.1947  &   & 0.1926 &  &  \textbf{0.2268}  &  \\
    \cline{1-2}
    \multirow{2}*{Forward} & Fangzhou &  
      0.1598 &    & \textbf{0.2032} &  &  0.1809  &  &  0.1887  &  \\
   &  Face  & 0.1332  &    &  0.1357  &   &  0.1119  &  &  \textbf{0.1503}  &  \\
     
  \hline 
  
  \hline
  \end{tabular}
  
\end{table}

 \begin{figure}[t]
\centering
\vspace{-0.2cm}
   \includegraphics[width=.85\linewidth]{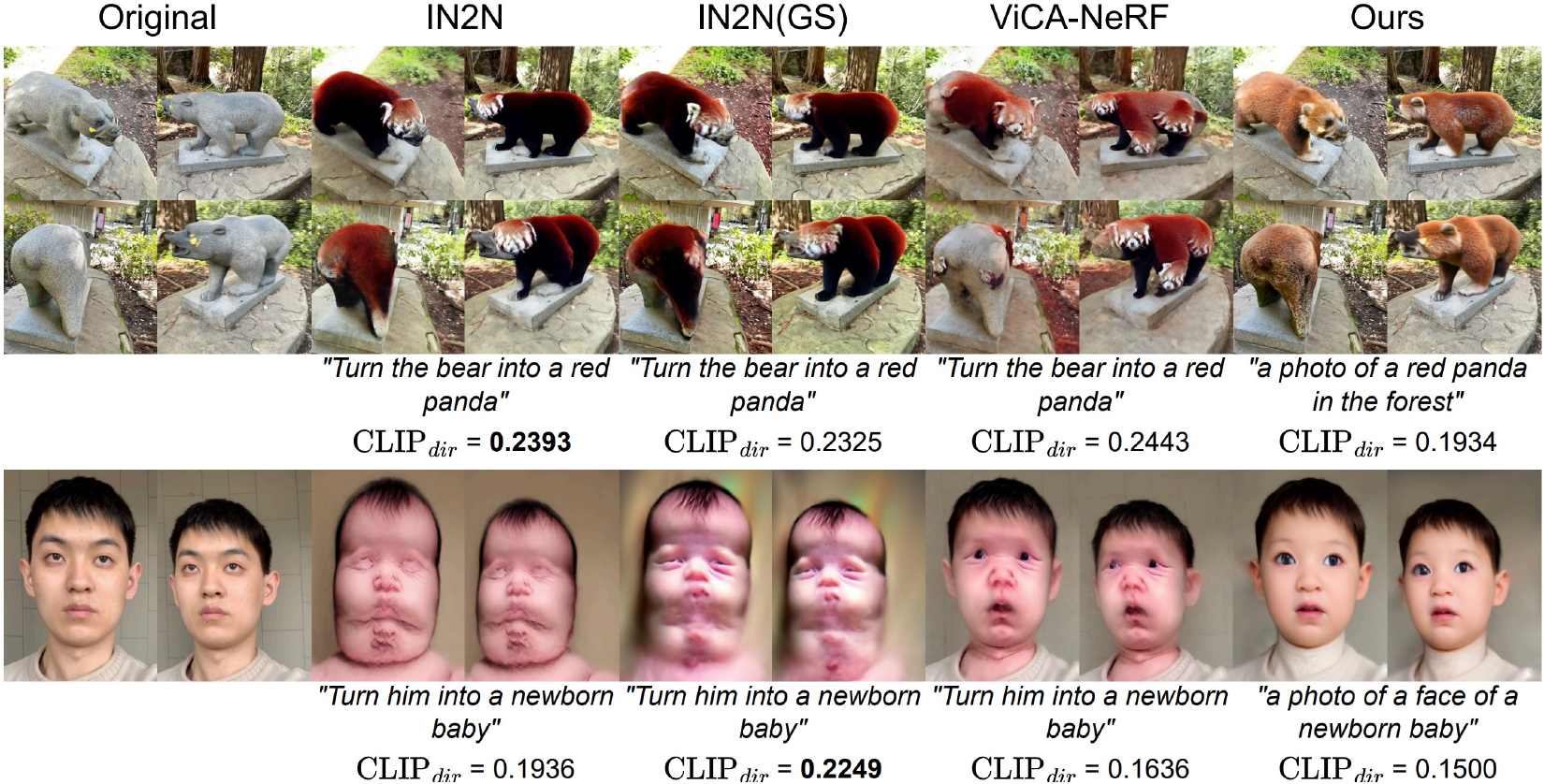}
    \caption{Failure cases of $\text{CLIP}_{dir}$. $\text{CLIP}_{dir}$ reflects the alignment between text instructions and editing results but ignores the editing quality. In the top row, our red panda has a lighter colour, making it score lower. Other methods have wrong face geometry but score higher than ours. In the bottom row, previous methods are closer to \textit{newborn baby}, making them score higher. However, the first two methods have terrible results, and ViCA's result is unnatural in the baby's eye areas. 
    } 
    \label{fig: failure clip}
    \vspace{-0.5cm}
\end{figure}

\subsection{Quantitative Evaluation}
We calculate the average $\text{CLIP}_{dir}$ over text instructions for each scene and summarize the results in \tabref{tab:quantitative eval}.
Our method outperforms other approaches in four out of six scenes.
However, we notice that $\text{CLIP}_{dir}$ may not always reflect the editing quality, as it measures more about the global similarity between the text prompt and the edited images, ignoring the majority of local details. We illustrate two failure cases of this metric in \figref{fig: failure clip}.
Our method generates better visual results but gets lower scores than previous methods. Therefore, we include as many examples as we can in the paper and the supplementary material to reflect the true visual quality of our method. We additionally provide the editing time for each method, and our method is the fastest among them. 


\subsection{Ablation Study}
\begin{figure}[t]
    \centering
    \vspace{-0.2cm}
   \includegraphics[width=.85\linewidth]{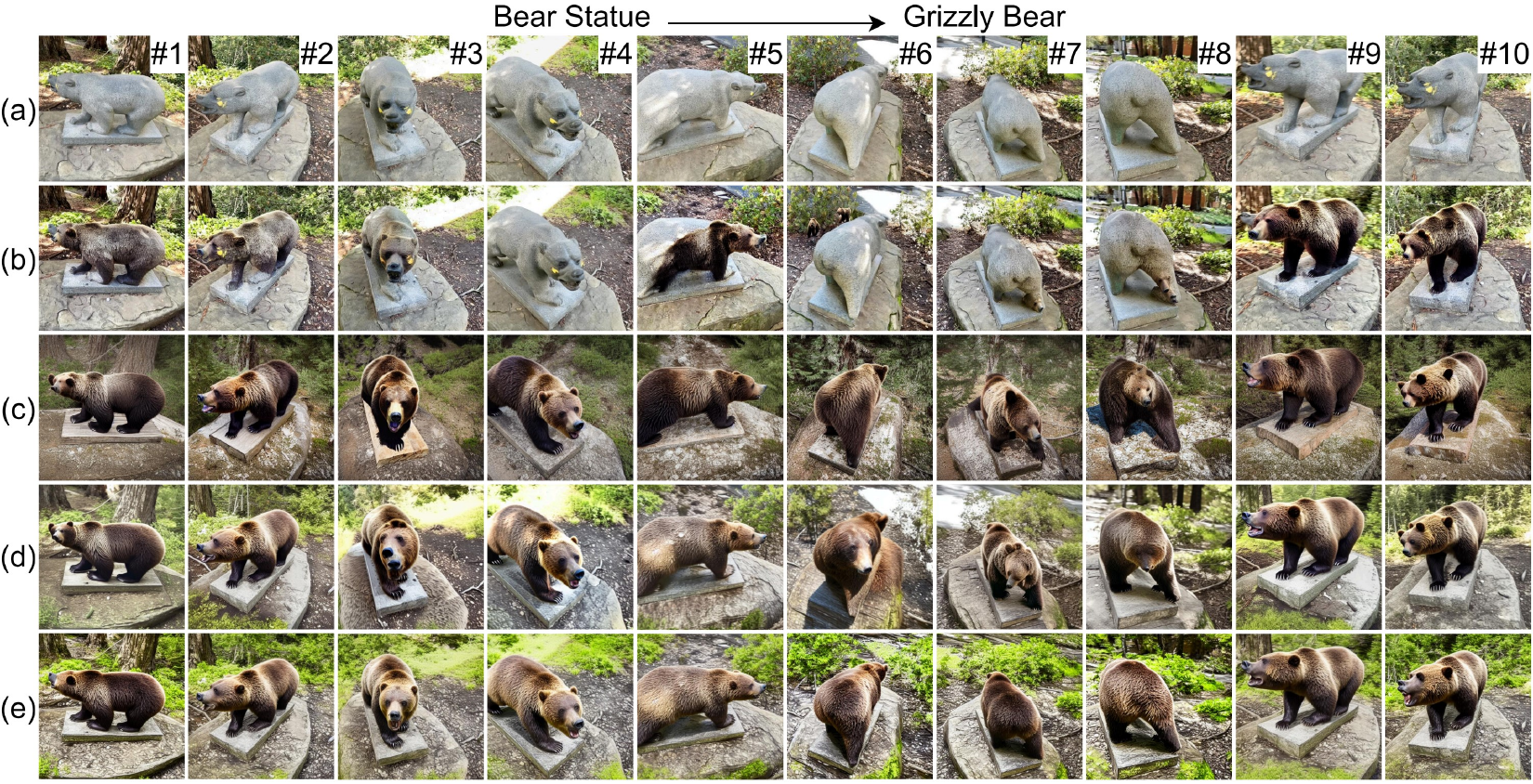}
    \caption{Ablation studies on the consistent editing. (a): Sampled images from the original dataset. 
    (b): Editing results using Instruct Pix2Pix~\cite{brooks2022instructpix2pix}. 
    (c): Our proposed depth-conditioned editing, which uses ControlNet with the randomly initialised latent codes. (d): Consistent initial latent code is applied by using DDIM inversion. (e): Attention-based latent code alignment is added based on (d).
    }
    \label{fig:ablation}
    \vspace{-0.5cm}
\end{figure}

We conduct ablation studies on our method to demonstrate the effectiveness of each proposed component. 
We selected the scene ``bear statue'', a 360-degree scene, as our subject because the 360-degree scene can better illustrate the effect of multi-view consistency. 
To ensure a fair comparison, the guidance scale $\lambda$ is set to 7.5 in all cases. 
Lang-SAM \cite{zhang2022glipv2, li2022elevater, li2021grounded} is not applied either to highlight the effect of each component on the scene's environment. 
The result is provided in \figref{fig:ablation}, where we show the original images in (a).

\noindent \textbf{One-time Instruct Pix2Pix Edit} (b): 
The most significant limitation of Instruct Pix2Pix (IPix2Pix) is that it fails to edit images in challenging viewpoints. 
For example, IPix2Pix fails to produce noticeable effects in challenging views \#4, 6, 7, and 8, where the bear statue is viewed from behind, and only partially alters views \#2, 3, and 5. 
This limitation prompts IN2N to choose iterative editing over one-time editing for the unstable performance of IPix2Pix. 
Even in relatively simpler views \#9,10, IPix2Pix still exhibits artefacts around the face of the bear, limiting the performance of IN2N and VICA-NeRF.

\noindent \textbf{ControlNet with Random Noise} (c): 
When ControlNet operates with random noise instead of inverted latent codes, it performs a generation task instead of an editing task.
While the generated images maintain consistency in geometry, owing to the incorporation of depth maps, their overall style diverges significantly from that of the original images.
Moreover, at challenging viewpoints such as views \#6,7,8, ControlNet forcefully generates front-facing views of a bear with the geometry of the bottom of the bear, which damages the eventual quality of 3D editing. 
Additionally, similar to IPix2Pix, it also suffers from artefacts when editing local details, such as the face area in view \#10. 

\noindent \textbf{ControlNet with Inverted Latent Codes (w/o AttnAlign)} (d):
When employing latent codes inverted from the original images, we observe a significant improvement in the general style alignment compared to using random noise. 
Additionally, texture and colour consistency are notably enhanced.
We attribute these enhancements to consistent initial latent codes brought by latent inversion. 
However, artefacts experienced by random noise still exist, such as the forceful front-facing views of the bear in view \#6,7,8 and artefacts around the facial area in view \#10. 

\noindent \textbf{ControlNet with Inverted Latent Codes  \& AttnAlign} (e):
After applying our proposed latent code alignment module, the artefacts presented in (c) and (d) are notably mitigated.  
By conditioning the editing on reference views using cross-view attention, the model searches for extra information about the to-be-edited image from the reference views. 
Therefore, it has a better understanding of the semantics and geometry of the image, avoiding producing forceful results. Such conditioning also unifies the appearances of edited images to a common ground, which further improves the quality of the final 3D editing. 

 \begin{figure}[t]
\centering
   \includegraphics[width=.95\linewidth]{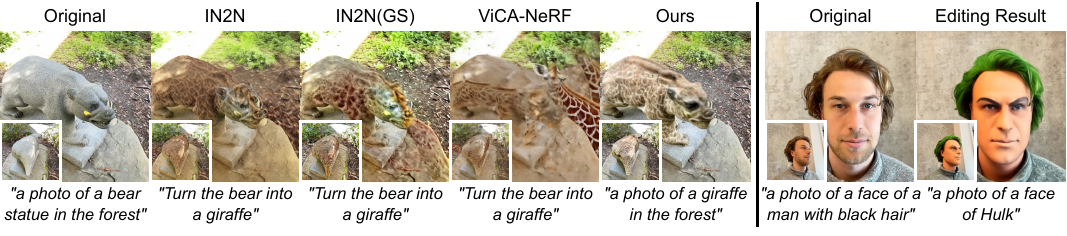}
   
    \caption{Failure cases. Left: Due to using depth guidance, our method cannot work well when a significant geometry change is required. However, we find that existing methods also cannot work well in this scenario even though they do not use depths. Right: our method fails when the 2D pre-trained diffusion model doesn't work well. Nevertheless, it shows that our method can still preserve the consistency.
    } 
    \label{fig:failure_case}
    \vspace{-0.5cm}
\end{figure}

\subsection{Limitations}
Some may be concerned about its ability to alter the scene's original geometry as we condition the editing on depth maps.
However, We find that in the most of 3D editing literature \cite{instructnerf2023, dong2023vica, zhuang2023dreameditor}, significant changes to the original geometry are not typically required. Instead, editing tasks often involve adjusting object styles, modifying background environments, or adding localized features, such as adding a moustache to a person. 
For the completeness of the experiments, we include an example that requires geometric changes. 
As illustrated on the left of \cref{fig:failure_case}, we fail to turn the bear statue into a giraffe. 
However, the same failure is also observed in IPix2Pix and baseline methods like IN2N and VICA-NeRF.
Another limitation is that the final result is not always faithful to the user's intention. 
As demonstrated in the right of \cref{fig:failure_case}, our method fails to transform the man into the comic character Hulk. 
We suspect the root of this problem lies in the ControlNet, which does not recognise the word ``Hulk", and does not produce the correct result. However, the consistency and sharp results demonstrate the effectiveness of our method. 

%% file: 05_conclusion.tex
\section{Conclusion}
In this paper, we propose an efficient 3D-aware consistency control editing method, \name, which greatly mitigates the artefacts and blurry results caused by the inconsistency in 2D editing, especially in 360-degree scenes. Based on a pre-captured Gaussian model, our method controls multi-view consistency by encouraging a consistency in all the stages of editing, \ie, Depth-conditioned Image Editing, and Attention-based Latent Code Alignment. We evaluate the performance of \name on diverse scenes, text prompts, and objects. Our method outperforms other state-of-the-art methods through our experiments. 

\noindent \paragraph{\textbf{Broader Impact:} }
Our method is one of the 3D editing methods that can be potentially misused for creating deceptive or harmful content, which could erode trust in digital media and exacerbate issues of misinformation and cyberbullying. 
By generating hyper-realistic alterations to images, videos, or even deepfakes, 3D editing technologies can be utilised to fabricate events, impersonate individuals, or manipulate scenes in ways nearly indistinguishable from reality. 
This capability not only leads to higher chances of confusion and misinformation but also opens pathways for harassment and defamation. 
Hence, it is necessary to enhance regulatory frameworks to mitigate these societal risks.